\documentclass[letterpaper]{article} 
\usepackage{aaai2026}  
\usepackage{times}  
\usepackage{helvet}  
\usepackage{courier}  
\usepackage[hyphens]{url}  
\usepackage{graphicx} 
\usepackage{natbib}  
\usepackage{caption} 
\usepackage{listings}
\usepackage[most]{tcolorbox}
\usepackage{xcolor}
\usepackage{algorithm}
\usepackage{algorithmic}
\usepackage{newfloat}
\usepackage{listings}
\DeclareCaptionStyle{ruled}{labelfont=normalfont,labelsep=colon,strut=off} 
\floatstyle{ruled}
\newfloat{listing}{tb}{lst}{}
\floatname{listing}{Listing}
\title{Fact4ac at the Financial Misinformation Detection Challenge Task: Reference-Free Financial Misinformation Detection via Fine-Tuning and Few-Shot Prompting of Large Language Models}
\author {
    Cuong Hoang\textsuperscript{\rm 1},
    Le-Minh Nguyen\textsuperscript{\rm 1}
}
\affiliations {
    \textsuperscript{\rm 1}Japan Advanced Institute of Science and Technology\\
    cuonghoang@jaist.ac.jp, nguyenml@jaist.ac.jp
}

\usepackage{bibentry}

\newtcolorbox{promptbox}[1][]{
    colback=gray!5, 
    colframe=gray!50, 
    fonttitle=\bfseries,
    coltitle=black,
    title=Prompt Instruction: #1,
    enhanced,
    attach title to upper,
    after title={\par\smallskip\hrule\smallskip},
    left=5pt,
    right=5pt,
    top=5pt,
    bottom=5pt,
    arc=2pt,
    boxrule=0.5pt,
}

\begin{document}

\maketitle

\begin{abstract}
The proliferation of financial misinformation poses a severe threat to market stability and investor trust, misleading market behavior and creating critical information asymmetry. Detecting such misleading narratives is inherently challenging, particularly in real-world scenarios where external evidence or supplementary references for cross-verification are strictly unavailable. This paper presents our winning methodology for the "Reference-Free Financial Misinformation Detection" shared task. Built upon the recently proposed RFC-BENCH framework \cite{jiang2026glistersgoldbenchmarkreferencefree}, this task challenges models to determine the veracity of financial claims by relying solely on internal semantic understanding and contextual consistency, rather than external fact-checking. To address this formidable evaluation setup, we propose a comprehensive framework that capitalizes on the reasoning capabilities of state-of-the-art Large Language Models (LLMs). Our approach systematically integrates in-context learning—specifically zero-shot and few-shot prompting strategies—with Parameter-Efficient Fine-Tuning (PEFT) via Low-Rank Adaptation (LoRA) to optimally align the models with the subtle linguistic cues of financial manipulation. Our proposed system demonstrated superior efficacy, successfully securing the first-place ranking on both official leaderboards. Specifically, we achieved an accuracy of \textbf{95.4\%} on the public test set and \textbf{96.3\%} on the private test set, highlighting the robustness of our method and contributing to the acceleration of context-aware misinformation detection in financial Natural Language Processing. Our models (14B and 32B) are available at \url{https://huggingface.co/KaiNKaiho}.
\end{abstract}


\section{Introduction}
\begin{figure*}[ht!]
    \centering
    \includegraphics[width=.8\linewidth]{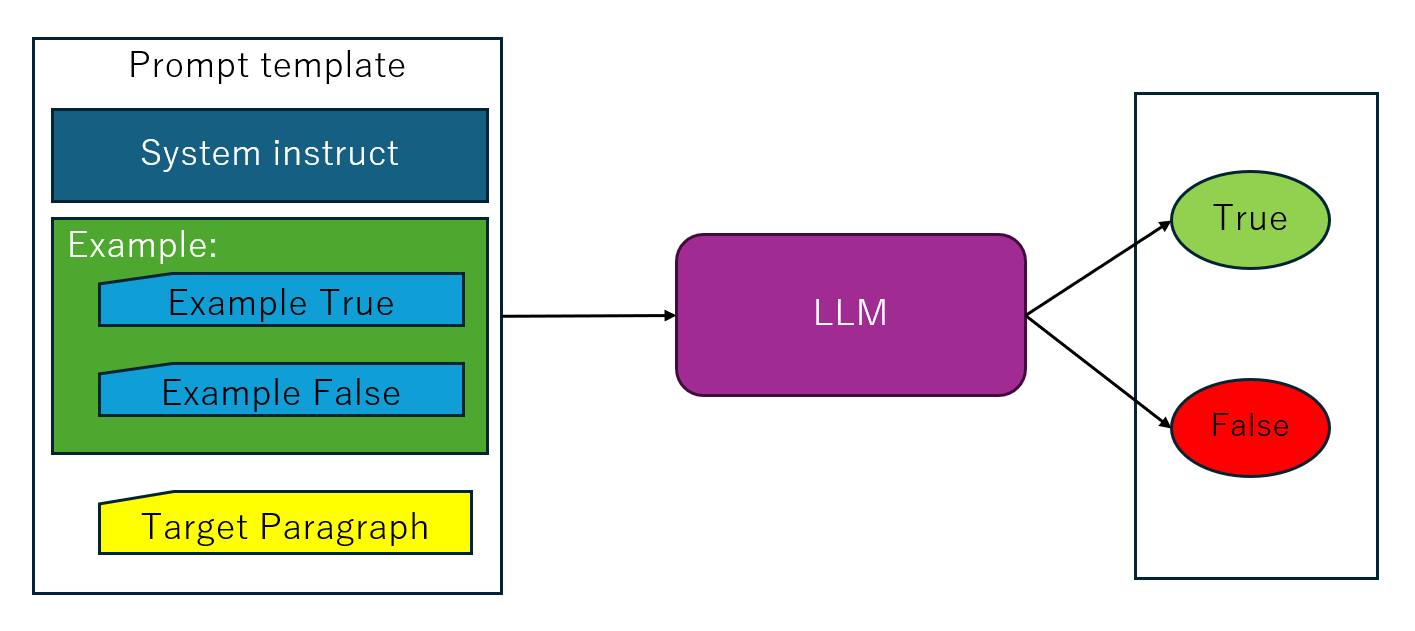}
    \caption{Our proposed approach is executed through a systematic pipeline combining in-context learning with parameter-efficient fine-tuning. Initially, we formulated a foundational zero-shot prompt to establish the base instruction. To enhance the model's contextual awareness, we empirically selected two representative exemplars—specifically, one instance from the 'True' class and one from the 'False' class—to serve as explicit demonstrations. These exemplars were then integrated with the base instruction to construct a unified few-shot prompt template. Subsequently, we applied Parameter-Efficient Fine-Tuning (PEFT) via Low-Rank Adaptation (LoRA) to train the model on this standardized template. This targeted tuning strategy facilitates the model's ability to deduce and internalize complex underlying patterns from the data, thereby optimizing its classification capabilities.}
    \label{fig:task2_mmfc_framework}
\end{figure*}
In the highly dynamic domain of financial markets, the integrity and completeness of news and analytical reports are of paramount importance. The deliberate or inadvertent omission of critical information in these texts can create severe information asymmetry, thereby misleading and manipulating investors. Consequently, this phenomenon not only exacerbates market volatility but also profoundly undermines public confidence in the broader financial ecosystem.

Detecting such missing information presents a formidable challenge in Natural Language Processing (NLP). Unlike traditional fact-checking tasks, identifying omissions is inherently complex due to the strict absence of supplementary evidence or explicit reference data for cross-verification. To bridge this critical gap, \cite{jiang2026glistersgoldbenchmarkreferencefree} introduced a novel, reference-free dataset that serves as a pioneering benchmark. This dataset establishes a crucial foundation, empowering researchers to leverage advanced computational methodologies—specifically Large Language Models (LLMs) and sophisticated NLP techniques—to systematically address this intricate problem. 

Formally, the task can be defined as a binary classification problem. Let $\mathcal{P}$ denote the space of all possible input paragraphs. Our proposed method acts as a classification function $S$ that maps an input paragraph $P \in \mathcal{P}$ to a predefined discrete label space. This relationship is formulated as:$$S: \mathcal{P} \rightarrow \{\text{True}, \text{False}\}$$Thus, for any given input paragraph $P$, the system outputs a prediction:$$S(P) \in \{\text{True}, \text{False}\}$$

Participating in this shared task, we propose a comprehensive framework that capitalizes on the reasoning capabilities of state-of-the-art LLMs. Our methodology encompasses a systematic exploration of zero-shot and few-shot prompting strategies, which is further enhanced by Parameter-Efficient Fine-Tuning (PEFT) utilizing Low-Rank Adaptation (LoRA) to optimally adapt the models to the nuances of the financial domain. Our proposed approach demonstrated superior efficacy, successfully securing the first-place ranking on both official leaderboards. Specifically, our system achieved an accuracy of 95.4\% on the public test set and an impressive 96.3\% on the private test set, highlighting its robustness and generalization capabilities in detecting unreferenced financial omissions.

\section{Related work}
In the foundational study introducing this dataset, the authors conducted a comprehensive evaluation of various prompting paradigms—namely, zero-shot, few-shot, and pairwise settings—to assess the baseline efficacy of state-of-the-art pretrained models. Their empirical analysis revealed that under zero-shot conditions, models exhibited substantial difficulty in accurately discriminating between factual and manipulated narratives. This limitation is primarily attributed to the inherent absence of supplementary context or external knowledge bases required for cross-verification. Conversely, the introduction of few-shot prompting (specifically in 2-shot and 8-shot configurations) yielded measurable improvements. Most notably, GPT5-Mini demonstrated the most significant performance leap compared to its zero-shot baseline, while models within the Qwen3 versions (encompassing both Thinking and Non-thinking variants) recorded an accuracy increase ranging from 2\% to 5\%.

However, these few-shot enhancements remain relatively marginal when juxtaposed with the results obtained via the pairwise evaluation strategy. Under the pairwise setting, the Qwen3, GPT, and DeepSeek model families exhibited remarkable average accuracy performance surges of 86.28\%, 97.13\%, and 90.55\%, respectively. This substantial disparity underscores a critical insight: when models are supplied with comparative reference information or explicit examples illustrating the structural transformation from a 'True' to a 'False' paragraph, their classification capabilities are profoundly amplified compared to scenarios where they are tasked with isolated, single-paragraph inference.

Further supporting the efficacy of example-driven learning, \cite{chen2022metalearninglanguagemodelincontext} proposed the framework of In-context Tuning for Language Models. Their research elucidates a promising trajectory: Large Language Models possess the intrinsic capacity to deduce underlying patterns and semantic nuances directly from strategically formulated examples within the prompt. This mechanism enables the models to accurately assign binary labels to unseen input paragraphs based on recognized patterns rather than external retrieval. This paradigm aligns perfectly with the unique reference-free constraints and objectives of the present shared task.

While massive, next-generation architectures (such as the Qwen3 version) have demonstrated commendable zero-shot capabilities, their practical deployment is heavily constrained by exorbitant computational resource requirements. Drawing upon empirical insights from previous literature \cite{10.1145/3746027.3762061,11309558}, we postulate a cost-effective yet highly performant alternative. Given that the foundational benchmark did not include the Qwen-2.5 architecture in its evaluations, we adopt this comparatively compact, highly optimized model to investigate whether the study's empirical findings generalize to this specific lineage \cite{qwen2025qwen25technicalreport,brown2020languagemodelsfewshotlearners}. By subjecting Qwen-2.5 to rigorous, domain-specific fine-tuning on this financial corpus, we aim to validate its diagnostic efficacy. Consequently, this approach not only provides deeper insights into the model's reasoning capabilities but also significantly mitigates the computational overhead and latency associated with real-world financial NLP applications.

\section{Methodology}
During our preliminary exploratory data analysis, a critical structural challenge emerged: distinguishing between 'True' and 'False' paragraphs is fundamentally intractable without prior exposure to the authentic counterpart. This limitation is starkly illustrated in the following example:
\begin{quote}
True: \\
"Hilton Worldwide (HLT) Up \underline{\textbf{11.2\%}} Since Last Earnings Report: Can It Continue?
Hilton Worldwide (HLT) reported earnings 30 days ago. What's next for the stock? We take a look at earnings estimates for some clues."

False: \\
"Hilton Worldwide (HLT) Up \underline{\textbf{0.6\%}} Since Last Earnings Report: Can It Continue?
Hilton Worldwide (HLT) reported earnings 30 days ago. What's next for the stock? We take a look at earnings estimates for some clues."
\end{quote}
As evidenced above, the manipulation relies on subtle numeric alterations rather than overt syntactic anomalies. This empirical observation, corroborated by the suboptimal zero-shot and few-shot performances (as detailed in Fig.\ref{tab_dev_result}) of pretrained LLMs, alongside the findings regarding pairwise evaluation from \cite{jiang2026glistersgoldbenchmarkreferencefree}, leads to a definitive conclusion: attempting to logically parse these texts based solely on internal reasoning—devoid of external factual retrieval or specific intrinsic knowledge—is an inherently flawed approach.


Consequently, we leverage Parameter-Efficient Fine-Tuning (PEFT) via LoRA to implicitly train the model to detect subtle linguistic cues of manipulation—such as flawed causal reasoning or sentiment misalignments—rather than memorizing discrete financial facts. This enables the model to infer text veracity autonomously without requiring external reference contexts. Furthermore, guided by the benchmark evaluations in \cite{jiang2026glistersgoldbenchmarkreferencefree}, we restrict our prompt engineering to a 2-shot configuration. Because scaling to an 8-shot setting yields negligible gains for sub-32B Qwen3 architectures, this optimal heuristic significantly reduces the computational footprint and context window demands without sacrificing overall efficacy.
\subsection{Data Preparation}


The shared task organizers provided two distinct datasets, both characterized by a perfectly balanced binary class distribution consisting of equal proportions ($50/50$) of 'True' and 'False' labels. For our initial experimental setup, we randomly partitioned the aggregated data into training, validation (dev), and testing subsets utilizing a $70:15:15$ split ratio. However, for final model training, we partitioned the aggregated dataset into an $80:20$ ratio and performed fine-tuning on the entire set. The rationale behind this strategy is discussed in further detail in "Result in Development set" section.
\subsection{Prompting}
Below are the templates for both the zero-shot and few-shot configurations, which are utilized during the fine-tuning process.
\begin{promptbox}[Zero-shot]
\small
\# Instruction \\
You are a financial misinformation detector. \\
Please check whether the following information is true or false and output the answer [true/false].

\medskip
\# Input \\
\textcolor{blue}{\{paragraph\_input\}} \\
The provided information is:
\end{promptbox}

\begin{promptbox}[Few-shot]
\small
\# Instruction \\
You are a financial misinformation detector. \\
Please check whether the following information is true or false and output the answer [true/false].

\medskip
\# Examples \\
\textcolor{blue}{\{example\_1\}}. The provided information is: \textcolor{blue}{\{label\_1\}}. \\
\textcolor{blue}{\{example\_2\}}. The provided information is: \textcolor{blue}{\{label\_2\}}. 

\medskip
\# Input \\
\textcolor{blue}{\{paragraph\_input\}} \\
The provided information is:
\end{promptbox}

\section{Result}

\subsection{Result in Development set}
\label{sec:dev_result}
\begin{table}[ht!]
    \centering
    \caption{Results of Development set.}
    \label{tab_dev_result}
    \resizebox{\linewidth}{!}{
    \begin{tabular}{llllll}
    \hline
    \textbf{Model name}    & \textbf{Accuracy}  & \textbf{Precision}  & \textbf{Recall}  & \textbf{F1}\\ \hline
    \multicolumn{5}{l}{\textbf{Pretrained}}  \\ \hline
    \multicolumn{5}{l}{Zero-shot} \\ \hline
    Qwen2.5-0.5B-Instruct  & 0.53  & 0.56 & 0.52 & 0.41       \\
    Qwen2.5-1.5B-Instruct  & 0.51   & 0.52  & 0.52  & 0.50        \\
    Qwen2.5-3B-Instruct  & 0.49   & 0.51  & 0.50  & 0.38        \\
    Qwen2.5-7B-Instruct  & 0.51   & 0.51  & 0.51  & 0.51           \\
    Qwen2.5-14B-Instruct  & 0.51   & 0.64  & 0.53  & 0.40        \\
    Qwen2.5-32B-Instruct  & 0.50   & 0.50  & 0.50  & 0.40        \\ \hline    
    \multicolumn{5}{l}{Few-shot}  \\ \hline
    Qwen2.5-0.5B-Instruct  & 0.51   & 0.26  & 0.50  & 0.34       \\
    Qwen2.5-1.5B-Instruct  & 0.56   & 0.56  & 0.56  & 0.56        \\
    Qwen2.5-3B-Instruct  & 0.50   & 0.75  & 0.52  & 0.36        \\
    Qwen2.5-7B-Instruct  & 0.53   & 0.53  & 0.53  & 0.53           \\
    Qwen2.5-14B-Instruct  & 0.50   & 0.54  & 0.52  & 0.46        \\
    Qwen2.5-32B-Instruct  & 0.53   & 0.54  & 0.53  & 0.52        \\ \hline    
    \multicolumn{5}{l}{\textbf{Fine-tuned}}  \\ \hline
    \multicolumn{5}{l}{Zero-shot} \\ \hline
    Qwen2.5-0.5B-Instruct  & 0.96    & 0.96  & 0.96  & 0.96      \\
    Qwen2.5-1.5B-Instruct  & 0.94  & 0.94  & 0.94  & 0.94        \\
    Qwen2.5-3B-Instruct  & 0.96  & 0.96  & 0.96  & 0.96         \\
    Qwen2.5-7B-Instruct  & 0.95   & 0.95  & 0.95  & 0.95           \\
    Qwen2.5-14B-Instruct  & 0.96   & 0.97  & 0.96   & 0.96        \\
    Qwen2.5-32B-Instruct  & 0.96   & 0.96  & 0.96  & 0.96        \\ \hline    
    \multicolumn{5}{l}{Few-shot}  \\ \hline
    Qwen2.5-0.5B-Instruct  & 0.86  & 0.87  & 0.86  & 0.86      \\
    Qwen2.5-1.5B-Instruct  & 0.89   & 0.89  & 0.89  & 0.89        \\
    Qwen2.5-3B-Instruct  & 0.92   & 0.93  & 0.92  & 0.92      \\
    Qwen2.5-7B-Instruct  & 0.89   & 0.89  & 0.89  & 0.89          \\
    Qwen2.5-14B-Instruct  & 0.96   & 0.97  & 0.97   & 0.96       \\
    Qwen2.5-32B-Instruct  & \underline{0.97}   & 0.97  & 0.97  & 0.97       \\ \hline         
    \end{tabular}
    }
\end{table}

As shown in Table \ref{tab_dev_result}, foundational pretrained models exhibit near-random performance (accuracy between 49\% and 56\%) regardless of scale or prompting strategy, which is consistent with the Qwen3 performance reported in \cite{jiang2026glistersgoldbenchmarkreferencefree}. While the 0.5B and 14B variants experienced a slight decline in accuracy, the remaining models demonstrated marginal gains of 1\% to 2\%, with the notable exception of the 1.5B version, which saw a 5\% increase. These findings further corroborate the experimental trends observed for Qwen3 in \cite{jiang2026glistersgoldbenchmarkreferencefree}.

Conversely, domain-specific fine-tuning yields a massive performance leap, resulting in an accuracy improvement of over 40\% compared to the pretrained baselines. However, a comparison between zero-shot and few-shot prompting reveals that only the 32B version benefited from the few-shot setting—achieving the highest overall accuracy of 97\%—while the performance of the other variants either stagnated or degraded. This suggests that fine-tuning with few-shot prompts may introduce bias into the models rather than facilitating the extraction of underlying data patterns.

Furthermore, a closer analysis of the misclassified instances highlights a shared characteristic: the complete absence of positive ('True') samples in the training set. This observation indicates that fine-tuning on the entire dataset would be substantially more effective than relying on the initial proportional splitting strategy.

\subsection{Result in Leaderboard}
\begin{table}[ht!]
    \centering
    \caption{Official Results at MisD@ICWSM2026 of Public and Private test set.}
    \label{tab_results_official}
    \resizebox{\linewidth}{!}{
    \begin{tabular}{llllll}
    \hline
    \textbf{Rank}   &\textbf{Team}    & \textbf{Accuracy}  & \textbf{Precision}  & \textbf{Recall}  & \textbf{F1}\\ \hline
    \multicolumn{6}{l}{\textbf{Public test}} \\ \hline
    \underline{1} &\underline{Fact4ac} \textbf{(our)}  & \underline{\textbf{0.954}}  & \underline{0.9567}  & \underline{0.9548}  & \underline{\textbf{0.954}}       \\
    2       & Coherence      & 0.9525   & 0.9542  & 0.9531  & 0.9524   \\
    3       & Alsper  & 0.9525   & 0.9549  & 0.9532  & 0.9524   \\  
    4   & DeepTruth  & 0.7469   & 0.7497  & 0.7459  & 0.7456   \\  
    5   & mfPE  & 0.6012   & 0.6012  & 0.6005  & 0.6001   \\  
    6    & baseline, 2 shot, GPT4.1  & 0.5521   & 0.5544  & 0.5534  & 0.5506   \\ \hline    
    \multicolumn{6}{l}{\textbf{Private test}}  \\ \hline
    \underline{1} &\underline{Fact4ac} \textbf{(our)}  & \underline{\textbf{0.963}}  & \underline{0.9654}  & \underline{0.9626}  & \underline{\textbf{0.9629}}       \\
    2       & Alsper      & 0.959   & 0.9619  & 0.9586  & 0.9589   \\
    3       & Coherence  & 0.955   & 0.9566  & 0.9547  & 0.9549   \\  
    4   & DeepTruth  & 0.755   & 0.7551  & 0.7551  & 0.7550   \\  
    5   & baseline, 2 shot, GPT4.1  & 0.570   & 0.5701  & 0.5694  & 0.5686   \\  
    6    & mfPE  & 0.555   & 0.5556  & 0.5554  & 0.5547   \\ \hline
    \end{tabular}
    }
\end{table}
The official results for both Public and Private test sets are summarized in Table \ref{tab_results_official}. Overall, our proposed method, Fact4ac, achieved the highest performance across all evaluation metrics, securing the first place in the competition.
\subsubsection{Comparative Performance on Public Test}
On the Public test set, our team demonstrated a robust capability in detecting misinformation with an Accuracy of 95.4\% and a dominant F1-score of 95.4\%. 
\begin{itemize}
    \item Comparison with Top Contenders: our team maintained a narrow but decisive lead over the two closest competitors, Coherence and AIsper, both of which shared an F1-score of 95.24\%.
    \item Gap with Baselines: Notably, our system significantly outperformed the GPT-4.1 (2-shot) baseline by nearly 40.34\% in F1-score, highlighting the limitations of general-purpose LLMs in specialized misinformation detection tasks without task-specific optimization.
\end{itemize}
\subsubsection{Superiority and Stability on Private Test}
The superiority of our team became even more evident on the Private test set, where the model's generalization ability was put to the test.
\begin{itemize}
    \item Consistent Improvement: our team not only retained its top position but also improved its performance compared to the Public set, reaching an Accuracy of 96.3\% and an F1-score of 96.29\%.
    \item Performance Leadership: Our system led the runner-up (Alsper) by 0.4\% and the third-place team (Coherence) by 0.8\% in F1-score. This gap, while seemingly small, is significant in high-performance competitive leaderboards, indicating a more precise calibration of our model's decision boundaries.
    \item Metric Balance: our team achieved the highest Precision (96.54\%) and Recall (96.26\%) simultaneously, proving that the system does not trade off between false positives and false negatives, but rather improves overall classification reliability.

\end{itemize}
\section{Discussion}
While the superior accuracy achieved by our proposed methodology across all evaluation sets validates its efficacy, we acknowledge certain limitations within our current framework. First, our empirical evaluations were constrained to the Qwen-2.5 architecture. Although this model family demonstrates robust domain adaptation, further investigation is required to ascertain whether these performance gains translate uniformly to other Large Language Model architectures. Second, formulating the objective as Sequence Classification inherently optimizes the model for a discrete binary output, effectively bypassing explicit intermediate reasoning pathways. This architectural choice, while highly efficient for this specific challenge, may constrain the framework's interpretability and generalizability to broader, reasoning-intensive tasks.

Crucially, a post-fine-tuning error analysis revealed a concentration of False Negatives (FN) specifically on entirely novel narratives—instances where neither the authentic nor the manipulated variants appeared in the training distribution. Rather than indicating an over-reliance on memorized data, this highlights a fundamental boundary of the reference-free paradigm: when financial manipulation relies on purely quantitative alterations (e.g., modifying a single numeric metric) without introducing structural flaws or semantic inconsistencies, the linguistic cues of deception are practically nonexistent. To overcome this limitation, our future work will focus on deepening the extraction of intrinsic, in-context linguistic anomalies. Furthermore, we intend to explore Knowledge Distillation paradigms. By leveraging high-capacity models (e.g., Qwen-32B) as 'Teachers' to generate explicit reasoning chains, we aim to distill this inferential capability into more compact, computationally efficient 'Student' Natural Language Inference (NLI) architectures.

Finally, we extend our profound gratitude to the organizers of the MisD@ICWSM2026 shared task for formulating such a novel and stimulating challenge. We also wish to acknowledge our fellow participating teams—Coherence, Aisper, DeepTruth, and mfPE—whose competitive spirit provided a dynamic learning environment that continuously inspired our methodological breakthroughs throughout the competition.

\bibliography{aaai2026}

\end{document}